\documentclass[sigconf]{acmart}

\AtBeginDocument{%
  \providecommand\BibTeX{{%
    \normalfont B\kern-0.5em{\scshape i\kern-0.25em b}\kern-0.8em\TeX}}}

\setcopyright{acmcopyright}

\copyrightyear{2020} 
\acmYear{2020} 
\setcopyright{acmcopyright}\acmConference[SIGIR '20]{Proceedings of the 43rd International ACM SIGIR Conference on Research and Development in Information Retrieval}{July 25--30, 2020}{Virtual Event, China}
\acmBooktitle{Proceedings of the 43rd International ACM SIGIR Conference on Research and Development in Information Retrieval (SIGIR '20), July 25--30, 2020, Virtual Event, China}
\acmPrice{15.00}
\acmDOI{10.1145/3397271.3401209}
\acmISBN{978-1-4503-8016-4/20/07}

\usepackage{latexsym}
\usepackage{algorithm}
\usepackage[noend]{algpseudocode}
\usepackage{url}
\usepackage{graphicx}
\usepackage{amsmath}
\usepackage{amsthm}
\usepackage{amssymb}

\newcommand{\pheadNoSpace}[1] {\noindent\textbf{#1.}} %
\newcommand{\pheadWithSpace}[1] {\vspace{1.25mm}\noindent\textbf{#1.}} %

\DeclareMathOperator*{\argmax}{argmax}

\begin{document}
\fancyhead{}
\title{Leveraging Adversarial Training in Self-Learning\\for Cross-Lingual Text Classification}

\author{Xin Dong$^1$, Yaxin Zhu$^1$, Yupeng Zhang$^1$, Zuohui Fu$^1$, Dongkuan Xu$^2$, Sen Yang$^3$, Gerard de Melo$^1$}
\renewcommand{\authors}{Xin Dong, Yaxin Zhu, Yupeng Zhang, Zuohui Fu, Dongkuan Xu, Sen Yang, Gerard de Melo}
\affiliation{\vspace{0.1cm}
\institution{$^1$ Rutgers University}
\institution{$^2$ The Pennsylvania State University}
\institution{$^3$ LinkedIn}
}
\email{{xd48, yz956, yupeng.zhang, zuohui.fu}@rutgers.edu, dux19@psu.edu, seyang@linkedin.com, gdm@demelo.org}

\renewcommand{\shortauthors}{Xin Dong, et al.}

\begin{abstract}
In cross-lingual text classification, one seeks to exploit labeled data from one language to train a text classification model that can then be applied to a completely different language. Recent multilingual representation models have made it much easier to achieve this. Still, there may still be subtle differences between languages that are neglected when doing so. To address this, we present a semi-supervised adversarial training process that minimizes the maximal loss for label-preserving input perturbations. The resulting model then serves as a teacher to induce labels for unlabeled target language samples that can be used during further adversarial training, allowing us to gradually adapt our model to the target language. Compared with a number of strong baselines, we observe significant gains in effectiveness on document and intent classification for a diverse set of languages.
\end{abstract}



\maketitle

\section{Introduction}

\pheadNoSpace{Background}
Text classification has become a fundamental building block in modern information systems, and there is an increasing need to be able to classify texts in a wide range of languages. However, as organizations target an increasing number of markets, it can be challenging to collect new task-specific training data for each new language that is to be supported.

To overcome this, cross-lingual systems rely on training data from a source language to train a model that can be applied to entirely different target languages \cite{deMeloSiersdorfer2007}, alleviating the training bottleneck issues for low-resource languages. 
Traditional cross-lingual text classification approaches have often relied on translation dictionaries, lexical knowledge graphs, or parallel corpora to find connections between words and phrases in different languages \cite{deMeloSiersdorfer2007}.
Recently, based on deep neural approaches such as BERT \cite{devlin2018bert}, there have been important advances in 
learning joint multilingual representations with self-supervised objectives \cite{devlin2018bert,artetxe2018massively,lample2019cross}. 
These have enabled substantial progress for cross-lingual training, by mapping textual inputs from different languages into a common vector representation space \cite{pires-etal-2019-multilingual}. With models such as Multilingual BERT \cite{devlin2018bert}, the obtained vector representations for English and Thai language documents, for instance, will be similar if they discuss similar matters.

Still, recent empirical studies \cite{singh-etal-2019-bert,libovick2019languageneutral} show that these representations do not bridge all differences between different languages. While it is possible to invoke multilingual encoders to train a model on English training data and then apply it to documents in a language such as Thai, the model may not work as well when applied to Thai document representations, since the latter are likely to diverge from the English representation distribution in subtle ways.

In this work, we propose a semi-supervised adversarial perturbation framework that encourages the model to be more robust towards such divergence and better adapt to the target language.
Adversarial training is a method to learn to resist small adversarial  perturbations that are added to the input so as to maximize the loss incurred by neural networks \cite{szegedy2013intriguing, goodfellow2014explaining}. %
Nevertheless, the gains observed from adversarial training in previous work have been limited, because it is merely invoked as a form of monolingual regularization.
Our results show that adversarial training is particularly fruitful in a cross-lingual framework that also exploits unlabeled data via self-learning.

\pheadWithSpace{Overview and Contributions}
Our model begins by learning just from available source language samples, drawing on a multilingual encoder with added adversarial perturbation. Without loss of generality, in the following, we assume English to be the source language.
After training on English, subsequently, we use the same model to make predictions on unlabeled non-English samples and a part of those samples with high confidence prediction scores are repurposed to serve as labeled examples for a next iteration of adversarial training until the model converges.

The adversarial perturbation improves robustness and generalization by regularizing our model. At the same time, because adversarial training makes tiny perturbations that barely affect the prediction result, the perturbations on words during self-learning can be viewed as inducing a form of code-switching, which replaces some original source language words with potential nearby non-English word representations.

Based on this combination of adversarial training and semi-supervised self-learning techniques, the model evolves to become more robust with regard to differences between languages.
We demonstrate the superiority of our framework on Multilingual Document Classification (MLDoc) \cite{schwenk2018corpus} in comparison with state-of-the-art baselines. Our study then proceeds to show that our method outperforms other methods on cross-lingual dialogue intent classification from English to Spanish and Thai \cite{schuster2018cross}. This shows that our semi-supervised adversarial framework is more effective than previous approaches at cross-lingual transfer for domain-specific tasks, based on a mix of labeled and unlabeled data via adversarial training on multilingual representations.

\section{Method}
\label{sec:method}

\begin{figure}[tbh]
    \centering
	\includegraphics[width=0.8\linewidth]{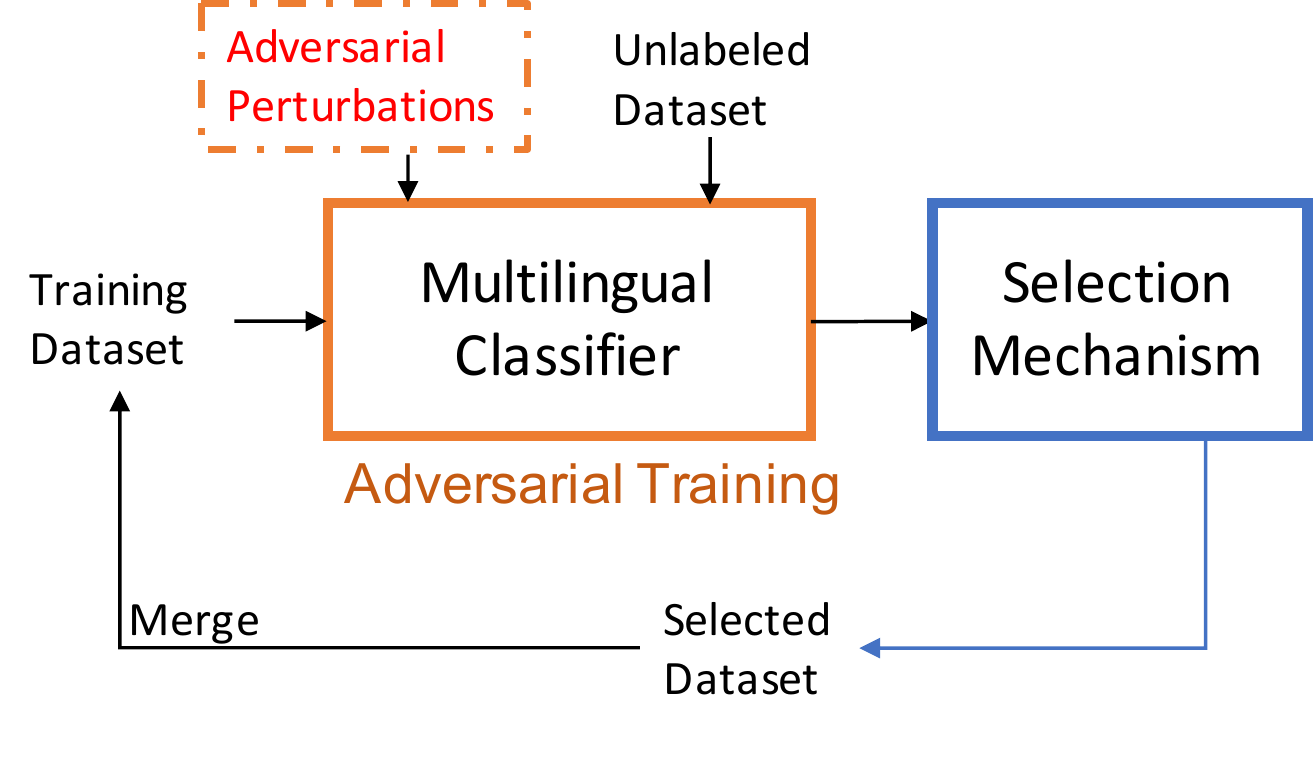}
	\caption{Illustration of self-learning process with adversarial training for cross-lingual classification.}
	\label{fig:sl}
\end{figure}

\pheadNoSpace{Overview of the Method}
Our proposed method consists of two parts, as illustrated in \autoref{fig:sl}.  The backbone is a multilingual classifier, which includes a pretrained multilingual encoder $f_{n}(\cdot; \theta_n)$ and a task-specific classification module $f_\mathrm{cl}(\cdot; \theta_\mathrm{cl})$.  By adopting an encoder that (to a certain degree) shares representations across languages, we obtain a universal text representation $\mathbf{h} \in \mathbb{R}^{d}$, where $d$ is the dimensionality of the text representation. The classification module $f_\mathrm{cl}(\cdot; \theta_\mathrm{cl})$ is applied for fine-tuning on top of the pretrained model $f_{n}(\cdot; \theta_n)$. It applies a linear function to map $\mathbf{h} \in \mathbb{R}^{d}$ into $\mathbb{R}^{|\mathcal{Y}|}$, and a softmax function, where $\mathcal{Y}$ is the set of target classes.

\pheadWithSpace{Adversarial Training}
Our adversarial self-learning process proceeds as follows.
First, we train the entire network $f(\cdot; \theta)$ in $K$ epochs using a set of labeled data $L = \{(\mathbf{x}_i, y_i) \mid i = 1,..., n\}$ from the source language, where $n$ is the number of labeled instances, $\mathbf{x}_i \in \mathcal{X}$ consists of embedding vectors $[\mathbf{v}_1, \mathbf{v}_2, ... , \mathbf{v}_T]$ for each instance ($T$ is the length of one sequence), and $y_i \in \mathcal{Y}$ are the corresponding ground truth labels. 

Adversarial training is motivated by the idea of making the model robust against adversarial examples. 
It is well-known that deep neural networks are susceptible to perturbed inputs that have been deliberately constructed to fool the network into making a misclassification \cite{szegedy2013intriguing}. Adversarial training is based on the notion of making the model robust against such perturbation, i.e., against an imagined adversary that seeks out minuscule changes to an input that lead to a misclassification, assuming that the class label should not actually be affected by such minuscule changes. 
To perform adversarial training, the loss function becomes:
\begin{gather}
    \label{loss_function}
    \mathcal{L}_{\mathrm{adv}}(\mathbf{x}_i, y_i)= \mathcal{L}(f(\mathbf{x}_i + \mathbf{r}_\mathrm{adv}; \theta), y_i) \\
    \text{where } \mathbf{r}_\mathrm{adv} = \argmax_{\mathbf{r}, ||\mathbf{r}||\leq\epsilon} \mathcal{L}(f(\mathbf{x}_i + \mathbf{r}; \Tilde{\theta}), y_i).  \nonumber
\end{gather}
Here $\mathbf{r}$ is a perturbation on the input and  $\Tilde{\theta}$ is a set of parameters set to match the current parameters of the entire network, but ensuring that gradient propagation only proceeds through the adversarial example construction process.
At each step of training, the worst case perturbations $\mathbf{r}_\mathrm{adv}$ are calculated against the current model $f(\mathbf{x}_i; \Tilde{\theta})$ in \autoref{loss_function}, and we train the model to be robust to such perturbations by minimizing \autoref{loss_function} with respect to $\theta$.
We later empirically confirm that adding random noise instead of seeking such adversarial worst case perturbations is not able to bring about similar gains (Section \ref{sec:results}). 

Generally, we cannot obtain a closed form for the exact perturbation $\mathbf{r}_\mathrm{adv}$, but 
Goodfellow et al. \cite{goodfellow2014explaining} 
proposed to approximate this worst case perturbation $\mathbf{r}_\mathrm{adv}$ by linearizing $f(\mathbf{x}_i; \Tilde{\theta})$ around $\mathbf{x}_i$.
With a linear approximation and an $L_2$ norm constraint in \autoref{pertubation}, the adversarial perturbation is
\begin{gather}
    \label{pertubation}
    \mathbf{r}_\mathrm{adv} \approx \epsilon \frac{\mathbf{g}}{||\mathbf{g}||_2} \\
    \text{where } \mathbf{g} = \nabla_{\mathbf{x}_i} \mathcal{L}(f(\mathbf{x}_i; \Tilde{\theta}))  \nonumber
\end{gather}
During the actual training, we optimize the loss function of the adversarial training in \autoref{loss_function} based on the adversarial perturbation defined by \autoref{pertubation} in each step.

\pheadWithSpace{Self-Learning}
Subsequently, in order to encourage the model to adapt specifically to the target language, 
the next step is to make predictions for the unlabeled instances in $U = \{\mathbf{x}_u \mid u = 1,..., m \}$. 
We can then incorporate unlabeled target language data with high classification confidence scores into the training set. 
To ensure robustness, we adopt a balanced selection mechanism, i.e., we first select a separate subset $\{\mathbf{x}_s \mid s = 1,...,K_\mathrm{t} \}$ of the unlabeled data for each class, consisting of the top $K_\mathrm{t}$ highest confidence items based on the current trained model.
The union set $U_\mathrm{s}$ of selected items is merged into the training set $L$ and then we retrain the model, again with adversarial perturbation. This process is repeated iteratively until some termination criterion is met.

\begin{table*}[tbh]
\centering
\small
\begin{tabular}{l c | c  c  c  c  c  c c}
\hline
Approach              & $en$&      $de$       &     $zh$       &      $es$    &      $fr$  &      $it$  &      $ja$  &      $ru$ \\ 
\hline
\emph{In-language supervised learning} & && && && &\\
~~ Schwenk et al.~(2018)~\cite{schwenk2018corpus} &  92.2 &93.7 &87.3 &94.5 &92.1 &85.6 &85.4 &85.7  \\
~~ BERT~(2018)~                    & 94.2  & 93.3   & 89.3& 95.7& 93.4 &88.0 &88.4 &87.5    \\
\hline
\emph{Cross-lingual transfer} & && && && &\\
--- \emph{Without unlabeled data}  & && && && &  \\
~~ Schwenk et al.~(2018)~\cite{schwenk2018corpus}&92.2&  81.2 &74.7& 72.5& 72.4&69.4&67.6& 60.8    \\
~~ Artetxe et al.~(2018)~\cite{artetxe2018massively}& 89.9  & 84.8  & 71.9  & 77.3 & 78.0 &69.4 &60.3&67.8\\

~~  BERT  & 93.0      &  75.0    & 71.3  & 78.3 &77.8&68.5 &71.8& 76.6\\
~~  BERT + Random Perturbation   & --      &  79.3   & 73.8  & 73.0 &81.3&67.1&73.1& 66.9\\ 
~~  BERT + Adv.\ Perturbation  & --      &  82.2   & 82.0  & 81.5 &83.7&72.9&75.6& 78.8\\
--- \emph{With unlabeled data}  & && && && &  \\
~~  Keung et al.~(2019)~\cite{keung2019adversarial} & --    &  88.1    & 84.7  & 80.8 &85.7&72.3&76.8& 77.4\\
~~  Dong et al.~(2019) \cite{dong2019robust} & --    &  89.9    & 84.5  & 84.8 &88.5&75.8&76.4& 79.3\\
~~  Our Approach w/ Random Perturbation & --    &  90.5    & 83.7   & 86.8 & 88.3 &76.1 &78.1 & 80.9\\
~~  Our Approach w/ Adv.\ Perturbation & --      &  \textbf{91.8}   & \textbf{86.7}  &\textbf{90.0}&\textbf{89.9}& \textbf{78.9}&\textbf{78.7}& \textbf{83.3}\\
\hline
\end{tabular}
\caption{Accuracy (in \%) on MLDoc experiments. Bold denotes the best on cross-lingual transfer.}
\label{tab:main-results}
\vspace{-20pt}
\end{table*}

\section{Experiments}
We evaluate our semi-supervised framework on two cross-lingual document and intent classification tasks to show the effectiveness of the combination of adversarial training and self-learning for Multilingual BERT-based cross-lingual transfer.

\subsection{Experimental Setup}
\pheadNoSpace{Datasets}  For evaluation, we first rely on MLDoc \cite{schwenk2018corpus}, a balanced subset of the Reuters corpus covering 8 languages for document classification, with 1,000 training and validation documents and 4,000 test documents for each language. The 4-way topic labeling consists of \emph{Corporate}, \emph{Economics}, \emph{Government}, and \emph{Market}. For cross-lingual classification, 1,000 target language training documents serve as unlabeled data for self-learning.

We further evaluate our framework on cross-lingual intent classification from English to Spanish and Thai \cite{schuster2018cross}. This dataset is built for multilingual task oriented dialogue. It contains 57k annotated utterances in English (43k), Spanish (8.6k), and Thai (5k) with 12 different intents across the domains \emph{weather}, \emph{alarms}, and \emph{reminders}. 3k Spanish or 2k Thai training utterances are used as unlabeled data for self-learning. All classification tasks are evaluated in terms of classification accuracy (ACC).

\begin{table}[]
    \centering
    \small
    \begin{tabular}{l|cc}
    \hline
    Parameter & MLDoc & CLIC \\
    \hline
    max.\ sequence length & 96 &32\\
    batch size & 64 & 128 \\
    learning rate & 2e-5 & 2e-6 \\
    $K_\mathrm{t}$ & 50& 30\\
    \# of training epochs & 5 & 6  \\
    $\epsilon$ & 1.0 & 1.0 \\
    \hline
    \end{tabular}
    \caption{ Hyper-parameters for our framework.}
    \label{tab:hyperparameter}
    \vspace{-25pt}
\end{table}

\pheadWithSpace{Model Details} 
We tune the hyper-parameters for our neural network architecture based on each non-English validation set. 
For the multilingual encoder, we invoke the Multilingual BERT model \cite{devlin2018bert}, which supports 104 languages\footnote{2018-11-23 version from \url{https://github.com/google-research/bert/blob/master/multilingual.md}}. 
Most hyper-parameters are shown in Table \ref{tab:hyperparameter}, with the exception that lower-casing is omitted for Thai and $\epsilon$ is 10 in the Japanese experiment. We rely on early stopping as a termination criterion, specifically, when the performance on the validation set stops improving in 2 self-learning iterations. 

\subsection{Results and Analysis}
\label{sec:results}
\pheadNoSpace{Cross-lingual Document Classification} Our MLDoc experiments compare our approach against several strong baselines. Schwenk et al.~\cite{schwenk2018corpus} propose MultiCCA, consisting of multilingual word embeddings and convolutional neural networks. Artetxe et al.~\cite{artetxe2018massively} pretrain a massively multilingual sequence-to-sequence neural MT model, invoking its encoder as a multilingual text representation used for fine-tuning on downstream tasks.
Keung et al.~\cite{keung2019adversarial} apply language-adversarial learning into Multilingual BERT during fine-tuning with unlabeled data. We also considered Multilingual BERT and itself with self-learning and adversarial training respectively as our baselines. Additionally, we compare Multilingual BERT with self-learning using unlabeled data as investigated by Dong et al.~\cite{dong2019robust}.
As shown in Table \ref{tab:main-results}, BERT with adversarial training outperforms Multilingual BERT across a range of languages, which establishes its merits for cross-lingual classification. Beyond this, our full framework further outperforms all baselines across 7 languages, including for phylogenetically unrelated languages.

\begin{table}[tbh]
\centering
\small
\begin{tabular}{lc|cc}
\hline
Approach              & $en$ & $es$ & $th$ \\ 
\hline
  Schuster et al.~(2018)~\cite{schuster2018cross} & 99.11 & 53.89 & 70.70\\
  Liu et al.~(2019)~\cite{liu2019zero}  & -- &90.20 & 73.43\\
  BERT  & 99.20  & 82.42  &62.77\\
  BERT + Adversarial Training  & 99.23 & 87.87& 67.20 \\
  BERT + Self-Learning  & -- & 88.33 & 71.51\\
  Our Approach   & -- & \textbf{92.41} & \textbf{75.95} \\
\hline
\end{tabular}
\caption{Accuracy (in \%) on cross-lingual intent classification without using labeled non-English data.}
\label{tab:intent}
\vspace{-25pt}
\end{table}
\pheadWithSpace{Cross-lingual Intent Classification} 
To evaluate the generalization of our framework to cross-lingual intent classification, we consider a diverse set of baselines as listed in Table \ref{tab:intent}. Schuster et al.~\cite{schuster2018cross} propose to combine Multilingual CoVe \cite{yu2018multilingual} with an auto-encoder objective and then use the encoder with a CRF model. We also run experiments on Multilingual BERT  and observe that it does not outperform the method from Liu et al.~\cite{liu2019zero}, because this method takes advantage of additional information by selecting 11 domain-related words as alignment seeds. However, our approach still achieves the new state-of-the-art result, which suggests that our adversarial framework for cross-lingual transfer is effective across different kinds of classification tasks.

\begin{table}[tbh]
\centering
\resizebox{1\linewidth}{!}{
\begin{tabular}{l | c  c  c c}
\hline
  & $en$ (0\%/0\%) & $de$ (16\%/52\% )   &     $zh$ (9\%32\%) &      $es$ (16\%/52\%)    \\\hline
BERT & 93.0 &  83.5   & 87.5 &  87.4   \\ 
+ Random Perturbation & 92.4  &  86.7   & 87.9  &  88.4      \\ 
+ Adv.\ Perturbation &\textbf{94.4}  &\textbf{88.8}    & \textbf{91.3} & \textbf{90.4} \\\hline
      &      $fr$ (15\%/50\%) &$it$ (14\%/49\%)  &      $ja$ (8\%/35\%)   &  $ru$ (12\%/44\%)\\\hline
 BERT      & 86.4 & 87.8 &  85.5  &84.0   \\ 
 + Random Perturbation  &87.8 &  89.1   & 86.3  &  85.5  \\ 
 + Adv.\ Perturbation  & \textbf{90.6} &\textbf{91.9}    & \textbf{88.6}  &  \textbf{89.2}\\\hline
\end{tabular}
}
\caption{Accuracy (in \%) on MLDoc English code-switching data. The respective ratios of replaced words from the vocabulary and replaced word token occurrences in the English test set are given in parentheses.}
\label{tab:code_switch}
\vspace{-35pt}
\end{table}

\pheadWithSpace{Influence of Adversarial Perturbations} 
To further evaluate the effect of adversarial perturbation and straight-forwardly show that it enables robustness with respect to divergence in the test set, we conduct an additional experiment on code-switching data. 
We create challenge datasets\footnote{Publicly available at \url{http://crosslingual.nlproc.org/}} that adopt the original English training data, while the test data consists of English documents in which we attempt to replace all vocabulary words with non-English translations based on the bilingual English to non-English dictionaries from MUSE\footnote{\url{https://github.com/facebookresearch/MUSE} -- We use a random but consistent choice in case there are multiple translations.}.
As a result, the test set documents consist of a form of code-switched language, in which many words are non-English but the word order remains unchanged. The replacement rates are listed in 
Table \ref{tab:code_switch}, along with the experimental results.
We observe that the baselines have a low accuracy when faced with such codeswitching in the test set. This applies to Multilingual BERT without perturbation as well as Multilingual BERT with random perturbation. 
In contrast, our adversarial perturbation is significantly more effective than no or random perturbation when dealing with this data, and it does not impede the accuracy on English compared with random noise, thus improving both generalization and robustness. 

\section{Related Work}
\label{sec:relatedwork}
\pheadNoSpace{Adversarial Training} There is substantial research on learning to resist adversarial perturbations with the goal of improving the robustness of a machine learning system \cite{szegedy2013intriguing, goodfellow2014explaining, madry2017towards}. In natural language processing, adversarial perturbation has proven effective for improving a model's generalization \cite{miyato2016adversarial,ebrahimi2017hotflip, cheng2019robust,zhu2019freelb}. Miyato et al.~\cite{miyato2016adversarial} adopt adversarial and virtual adversarial training for improved semi-supervised text classification in monolingual settings. 
Fu et al.~\cite{fu2020absent} propose an Adversarial Bi-directional Sentence Embedding Mapping framework to learn cross-lingual mappings of sentence representations. Cheng et al.~\cite{cheng2019robust} improve a translation model with adversarial source examples. In our experiments, we show that our approach outperforms adversarial perturbation applied to Multilingual BERT.

\pheadWithSpace{Cross-lingual Representation Learning} 
While cross-lingual text classification has a long history \cite{deMeloSiersdorfer2007,DongDeMelo2018CLSentimentEmbedding}, recent work in this area has been inspired by the success of deep neural models such as BERT \cite{devlin2018bert}. Multilingual extensions of such pretrained models include the multilingual version of BERT trained on the union of 104 different language editions of Wikipedia \cite{devlin2018bert}.
Artetxe et al.~\cite{artetxe2018massively} show that cross-lingual sentence embeddings can be obtained based on the encoder from a pretrained sequence-to-sequence model.
Lample et al.~\cite{lample2019cross} incorporate parallel text into BERT's architecture by training on a new supervised learning objective. Our framework can flexibly be used in conjunction with any of these methods, since they provide a multilingual representation space shared across languages. Our experiments show that our approach yields substantially better results than relying on such models directly.

\section{Conclusion}
\label{sec:conclusion}
While multilingual encoders have enabled better cross-lingual learning, the obtained models often are not attuned to the subtle differences that a model may encounter when fed with documents in an entirely new language. To adress this, this paper proposes an adversarial perturbation framework that makes the model more robust and enables an iterative self-learning process that allows the model to gradually adapt to the target language. 
We achieve new state-of-the-art results on cross-lingual document and intent classification and demonstrate that adversarial perturbation is an effective method for improved classification accuracy without any labeled training data in the target language. 

\bibliographystyle{ACM-Reference-Format}
\bibliography{acmart}

\end{document}